\documentclass[letterpaper, 10 pt, conference]{ieeeconf}  % Comment this line out if you need a4paper

\IEEEoverridecommandlockouts                              % This command is only needed if 
\usepackage{amsmath,amssymb,amsfonts}
\usepackage{graphicx}
\usepackage{hyperref}
\usepackage{booktabs}
\usepackage{tabularx}
\usepackage{subcaption}
\usepackage{siunitx}
\usepackage{makecell}

\title{\LARGE \bf
CLIPping the Limits: Finding the Sweet Spot for Relevant Images in Automated Driving Systems Perception Testing
}

\author{Philipp Rigoll$^{1}$, Laurenz Adolph$^{1}$ , Lennart Ries$^{1}$ and Eric Sax$^{2}$% <-this % stops a space
\thanks{$^{1}$Philipp Rigoll, Laurenz Adolph and Lennart Ries are with FZI Research Center for Information Technology, 76131 Karlsruhe, Germany
        {\tt\small philipp.rigoll@fzi.de, adolph@fzi.de, ries@fzi.de}}
\thanks{$^{2}$Eric Sax is with Karlsruhe Institute of Technology, 76131 Karlsruhe, Germany
        {\tt\small eric.sax@kit.edu}}%
}

\begin{document}

\maketitle
\thispagestyle{empty}
\pagestyle{empty}

%%%%%%%%%%%%%%%%%%%%%%%%%%%%%%%%%%%%%%%%%%%%%%%%%%%%%%%%%%%%%%%%%%%%%%%%%%%%%%%%
\begin{abstract}
Perception systems, especially cameras, are the eyes of automated driving systems.
Ensuring that they function reliably and robustly is therefore an important building block in the automation of vehicles.
There are various approaches to test the perception of automated driving systems.
Ultimately, however, it always comes down to the investigation of the behavior of perception systems under specific input data.
Camera images are a crucial part of the input data.
Image data sets are therefore collected for the testing of automated driving systems, but it is non-trivial to find specific images in these data sets.
Thanks to recent developments in neural networks, there are now methods for sorting the images in a data set according to their similarity to a prompt in natural language.
In order to further automate the provision of search results, we make a contribution by automating the threshold definition in these sorted results and returning only the images relevant to the prompt as a result.
Our focus is on preventing false positives and false negatives equally.
It is also important that our method is robust and in the case that our assumptions are not fulfilled, we provide a fallback solution.

\end{abstract}

%%%%%%%%%%%%%%%%%%%%%%%%%%%%%%%%%%%%%%%%%%%%%%%%%%%%%%%%%%%%%%%%%%%%%%%%%%%%%%%%
\section{INTRODUCTION}
Malfunctioning automated driving systems (ADS) carry the risk of causing damage and injuring or, in the worst case, killing people. 
ADS make their decisions based on their perception.
It is therefore of utmost importance to test the perception excessively and ensure its robustness.
We focus on image-based perception.
To test these perception systems, the corresponding test data sets must be composed.
This involves finding those images that are similar to images that previously showed abnormalities in tests.
As an example, we can assume that we develop an ADS and realize that situations with fog are a challenge for the system.
We also assume that the data is not expensively labeled and that we have no annotations regarding the weather.
For testing and development, it would be useful to have a data set consisting of images with fog.
The aim is therefore to identify the foggy images in our full data set and to create this partial data set.
However, this should result in as little additional manual work as possible and a largely automated solution is necessary in order not to negatively impact the development time of the ADS.

A semantic search in the images is necessary to identify the desired images in the full data set.
Since humans are good at expressing semantics in natural language, a search based on natural language is being investigated.
The actual search should then run as automatically as possible.
A state-of-the-art approach for such a search are neural networks that transform images and texts into a common latent space.
In this latent space, similar images and similar texts are assigned similar representations.
When a search is initiated, the text input is transferred by a neural network into its representation in latent space.
All images in the data set can then be sorted according to their distance from the input vector and thus their similarity to the text input.
A well-known, pre-trained neural network that enables such a search is CLIP (Contrastive Language-Image Pre-Training)~\cite{radford_learning_2021-1}.

The disadvantage of this procedure is that the images in the database are only sorted.
It is not determined up to which image the result matches the text request.
It is therefore still necessary to manually define a limit up to which of the sorted images the results are relevant.
In our example, this means that CLIP sorts the images so that the foggy images appear first in the result.
However, we do not know how many images are actually foggy and therefore cannot create a partial data set consisting of foggy images.
Thus, we have to manually check the search results to see how many images are actually foggy and define a limit.
This means that the process is not fully automated and the development of the ADS is delayed because we have to manually inspect the sorted images.

\begin{figure}%
	\centering
	\includegraphics[width=.95\columnwidth]{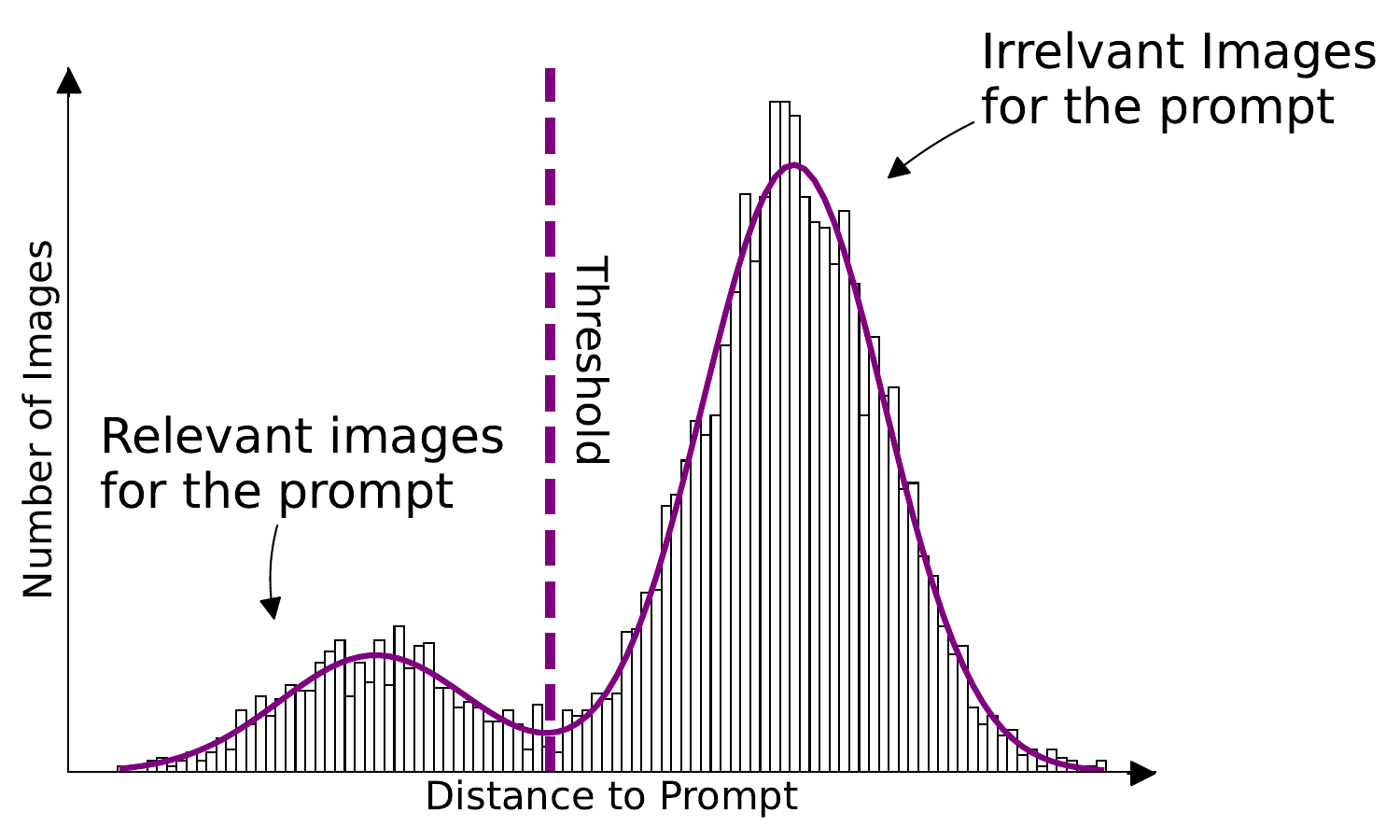}%
	\caption{Schematic illustration of our method for determining a threshold value based on the distribution of the similarity values.}%
	\label{fig:pull_image}%
\end{figure}

Therefore, we present a method for automatically finding a threshold value (see Fig.~\ref{fig:pull_image}).
This means that it is no longer necessary to look at the sorted images manually in order to set a threshold value.
As we will show later, a fixed threshold value is not possible because the range of the values of the similarity metric depends on the prompt.
We will give equal importance to the occurrence of false positives and false negatives.
Neither is ideal, but both are acceptable in the context under consideration.
False positives mean that data is used in the development that does not correspond to the request.
In our example, these would be images that are not foggy.
Our partial data set would therefore not consist exclusively of images with fog.
The test would therefore be more extensive than necessary.
However, this is not harmful as long as true positive results, i.e. images with fog, are included in the test.
False negatives mean that images that actually match the search query are not included in the test.
In our example case, our partial data set therefore does not contain all the foggy images that are present in the full data set.
However, a selection and reduction of data for testing is necessary in order to keep the testing effort within acceptable limits.
Both errors are already inherent in the CLIP sorting.
We are not able to prevent both errors simultaneously by determining the threshold value, but we minimize their combined effect by giving them the same weighting.

\section{RELATED WORK}
Xia et al.~\cite{xia_automated_2023} emphasize the importance of data and its compilation in the development of ADS.
King et al.~\cite{king_taxonomy_2020} describe how recorded data can be used for the validation of ADS.
They also describe that specified test cases must be examined.

Therefore, in the face of growing data sets~\cite{liu_survey_2024}, powerful search algorithms become necessary.
In the case of images, there are various approaches for searching.
One option are textual descriptions of the images.
This can be done manually~\cite{alkhawlani_text-based_2015} or with the help of neural networks~\cite{stefanini_show_2023}.
However, manual annotation is time-consuming, and it may not be clear in advance which image properties should be included in the annotation.
Another alternative is metadata, which contains structured information about the images~\cite{naito_browsing_2010,klitzke_real-world_2019,rigoll_scalable_2022}.
But even in this case, it is not clear which metadata will be necessary for future searches.

A generalized search method is therefore necessary.
Content-based search methods that look at higher-level features can be used for this purpose.
They try to close the semantic gap between images and texts and are expressive~\cite{lew_content-based_2006}.
One method that closes this gap is CLIP~\cite{radford_learning_2021-1}.
CLIP consists of an encoder for images and an encoder for text.
Both encoders map the input to the same latent space, whereby similar images and similar texts are mapped to similar vectors.
Radford et al.~\cite{radford_learning_2021-1} demonstrate the ability to close the semantic gaps using zero shot classification problems.
We have demonstrated that CLIP can be used for image retrieval in the automotive context~\cite{rigoll_focus_2023}.
The method has also been expanded to enable searching for individual object properties within automotive images~\cite{rigoll_unveiling_2023}.
However, the results only ever consist of the sorted data set, and it is not clear up to which image the data corresponds to the query.

\section{METHOD}
To determine a threshold value $\tau$ for the search, we start with an image data set~$\{I_i\}^n_{i=1}$.
The $k$-dimensional vector representation is calculated in advance for each of these images using the CLIP image encoder~$\mathcal{E}_\text{image}$:
\begin{equation}
    \mathbf{v}_i = \mathcal{E}_\text{image}(I_i) \in \mathbb{R}^k.
\end{equation}
The image retrieval is initiated by the user through a textual search prompt $T_\text{query}$ in natural language.
The text encoder $\mathcal{E}_\text{text}$ of CLIP is used to calculate the vector representation of this request $\mathbf{v}_\text{query}=\mathcal{E}_\text{text}(T_\text{query}) \in \mathbb{R}^k$.
The vector representations of the images in the database are then sorted according to their distance to this query vector.
The cosine distance is used as the distance measure, which is based on the cosine similarity~\cite{national_institute_of_standards_and_technology_nist_cosine_2023}.
The cosine distance to the request vector is therefore calculated for each vector representation of the images:
\begin{equation}
    d_\text{cos}(\mathbf{v}_\text{query}, \mathbf{v}_i) = 1 - \frac{\mathbf{v}_\text{query}^T \mathbf{v}_i}{\lVert\mathbf{v}_\text{query} \rVert \lVert \mathbf{v}_i\rVert}.
\end{equation}
The images with the smallest cosine distances should be most similar with the textual query.
Now a threshold value is to be defined for the cosine distance up to which it is assumed that the images correspond to the query:
\begin{equation}
\{ I_i | d_\text{cos}(\mathbf{v}_\text{query}, \mathbf{v}_i) \leq \tau \text{, for } i=1,2,\dots,n\}.
\end{equation}

For the following experiments, the ACDC data set~\cite{sakaridis_acdc_2021} is used.
The data set consists of $\SI{8012}{}$ images: $\SI{4006}{}$ images with clear sky at daytime, $\SI{1000}{}$ images with fog, $\SI{1006}{}$ images at nighttime, $\SI{1000}{}$ images with rain and $\SI{1000}{}$ images with snow.

\begin{figure}%
	\centering
	\includegraphics[width=.95\columnwidth]{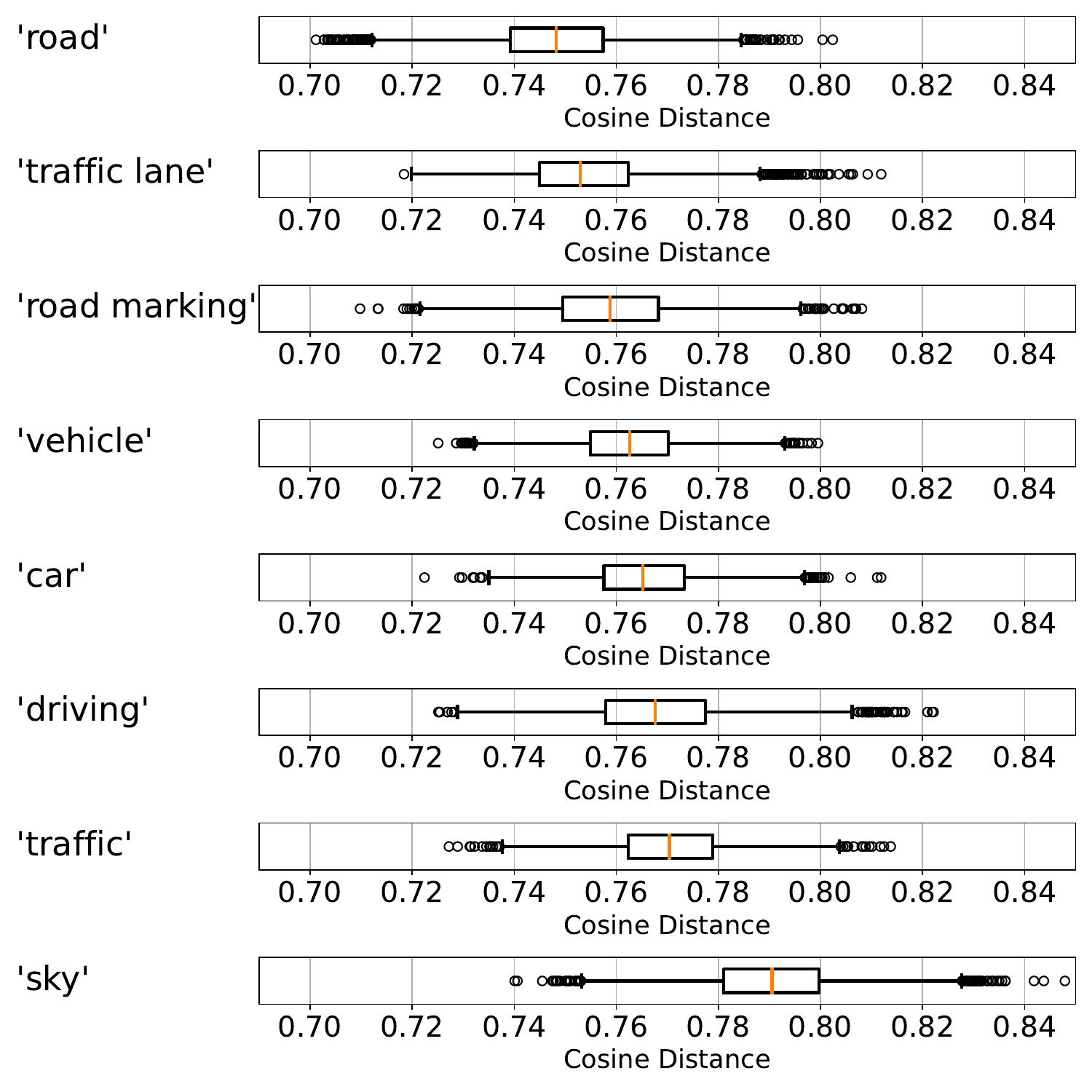}%
	\caption{Range of cosine distances of different prompts to all images in the ACDC~\cite{sakaridis_acdc_2021} data set}%
	\label{fig:range_prompts}%
\end{figure}

As the range of the cosine distance values depend on the corresponding prompt, it is not possible to select a fixed threshold value.
The range of the cosine distance values are even independent of the extent to which the images actually match the query.
In order to show that, we have chosen queries such as `road' and `sky', which are visible in many images, and looked at the range of the cosine distance values (see Fig.~\ref{fig:range_prompts}).
This unequal spectrum of distances was also observed by Wang et al.~\cite{wang_balanced_2024}.
They try to counter this by evaluating several auxiliary prompts.
For performance reasons, we base our approach on the results of a single query.

To define a threshold value, we model the distribution of the cosine distances as the sum of two Gaussian distributions:
\begin{equation}
    f_\Theta(x) = a_1 \cdot e^{-\frac{1}{2} \left(\frac{x-\mu_1}{\sigma_1} \right)^2} + a_2 \cdot e^{-\frac{1}{2} \left(\frac{x-\mu_2}{\sigma_2} \right)^2}.
\end{equation}
This is based on the assumption that one of the Gaussian distributions represents the images that match the query~$Q$ and the other the images~$\overline{Q}$ that do not match.
To determine the parameters of this function, the cosine distances are first converted into a normalized histogram.
We select the histogram so that the smallest bin starts at the smallest observed cosine distance and the largest bin ends at the largest observed cosine distance.
We chose 100 bins, each of which are the same size and normalize the histogram so that the integral over the range is~1.
To obtain the parameterization $\Theta = (a_1, \mu_1, \sigma_1, a_2, \mu_2, \sigma_2)$ of $f(x)$, the sum of the two Gaussian distributions is fitted to the histogram values.

We use a non-linear least squares method for the fitting process.
As initial parameters for this process we set the amplitudes $a_1, a_2$ to 1, the parameters $\mu_1, \mu_2$ to the mean of the cosine distances and the parameters $\sigma_1, \sigma_2$ to the standard deviation of the cosine distances.

However, the modeling with the sum of two Gaussian distributions only makes sense as long as there is a sufficient number of images that fall into both groups.
If only a few images fall into one of the two groups, a fallback method must be used.
We assume this to be the case if the non-linear least square method does not converge.
As a fallback, we additionally model the cosine distances as a single Gaussian distribution:
\begin{equation}
    g_\phi(x) = a_1 \cdot e^{-\frac{1}{2} \left(\frac{x-\mu_1}{\sigma_1} \right)^2},
\end{equation}
with the parameters $\phi = (a_1, \mu_1, \sigma_1)$ and try to fit it to the histogram data.
This approach is based on the assumption that the cosine distances within the images that are similar to the query~$Q$ and the cosine distances of the images that are dissimilar~$\overline{Q}$ to the query are normally distributed.
In addition, it is assumed that in each case the data of the other group~(depending on the case~$Q$ or~$\overline{Q}$) is relatively small and hardly influences the overall distribution.
If the fitting of the single Gaussian distribution does not converge either, no meaningful threshold value can be determined with our method and the results must be analyzed manually.
However, we did not observe this situation in our experiments.

\begin{figure}%
	\centering
	\includegraphics[width=.95\columnwidth]{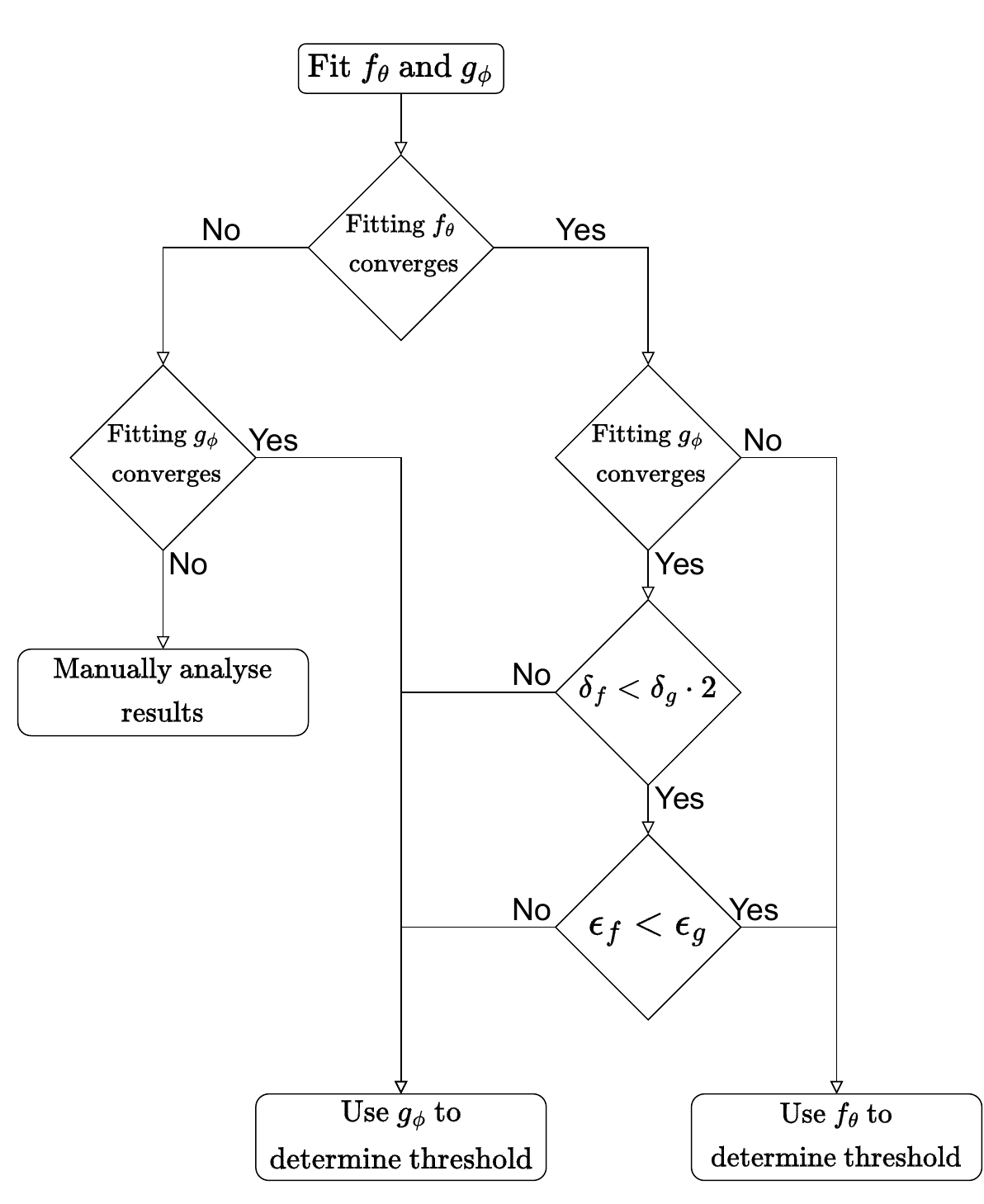}%
	\caption{Procedure to decide which modeling best fits the distribution of the cosine distances.}%
	\label{fig:decision_tree}%
\end{figure}

If both the sum of the two Gaussian distributions and the single Gaussian distribution can be fitted to the histogram data, it is checked which of the two functions describes the data better.
For this purpose, the uncertainty estimate of the parameters $\delta_f, \delta_g$ from the non-linear least squares method and the sum of the absolute errors $\epsilon_f, \epsilon_g$ are examined.
The uncertainty estimate of the parameters is calculated by the sum of the diagonal elements of the covariance matrix.
The procedure for deciding which modeling to use is shown in Fig.~\ref{fig:decision_tree}.
Since the modeling with the sum of two Gaussian distributions is preferred, an uncertainty estimate of the parameters up to twice as large is tolerated: $\delta_f < \delta_g \cdot 2$.
We have chosen the factor $2$ because $\delta_f$ has twice as many parameters as $\delta_g$.
However, the choice is more or less arbitrary, but has led to satisfactory results in our experiments.
Additionally, if the sum of the two Gaussian distributions is to be selected for modeling, the sum of the absolute errors between the fitted curve and the histogram data $\epsilon_f < \epsilon_g$ must also be smaller.

\subsection{Sum of two Gaussian distributions}
If the distribution of cosine distances is modeled by the sum of two Gaussian distributions, we assume that the Gaussian with the smaller mean value belongs to the distribution of images matching the query.
Since we have the parameterization of the individual Gaussian distributions, they can be described separately.
The two Gaussian distributions should be separated in such a way that neither false positives nor false negatives are favored.
The intersection of the two Gaussian distributions, which lies between the two mean values of the Gaussian distributions, is suitable for this purpose.
For this purpose, the zeros of the difference of the two Gaussian distributions are computed.
We are only interested in the intersection $\tau$ that lies between the two mean values:
\begin{align}
    a_1 \cdot e^{-\frac{1}{2} \left(\frac{\tau-\mu_1}{\sigma_1} \right)^2} &- a_2 \cdot e^{-\frac{1}{2} \left(\frac{\tau-\mu_2}{\sigma_2} \right)^2} = 0,\\
    (\mu_1 \le \tau \le \mu_2) &\vee (\mu_2 \le \tau \le \mu_1)
\end{align}
As a starting point for the search, we use: $\frac{\mu_1 + \mu_2}{2}$.
This intersection point is then used as the threshold value $\tau$ for the cosine distances.

\subsection{Fallback: single Gaussian distribution}
In the case that the sum of two Gaussian distributions does not allow a good modeling of the histogram values, a single Gaussian distribution is used as a fallback to determine the threshold value.
In this case, it is assumed that the outliers with small cosine distances match the search query.
There is no fixed definition for outliers in Gaussian distributions, which is why the threshold value must be estimated.
We have opted for deviations greater than two standard deviations:
\begin{equation}
    \tau = \mu_1 - 2 \cdot \sigma_1.
\end{equation}

\section{EVALUATION}
In our experiments, we use the ViT-B/32 architecture\footnote{\url{https://github.com/openai/CLIP}} of CLIP with a $k=512$ dimensional vector representation.
For the evaluation of our method, we use the optimum F1 score $F_{1\text{max}}$ as a reference value.
The F1 score weights the recall and precision equally, and the maximum value is~1:
\begin{equation}
    \frac{2 \cdot \text{\#true positives}}{2\cdot \text{\#true positives} + \text{\#false negatives} + \text{\#false positives}}.
\end{equation}
The F1 score can therefore only be calculated with labeled data.
To find the optimum F1 score, the cosine distance of every image is treated as a threshold candidate.
Using these potential thresholds, all F1 scores are calculated and the threshold with the highest F1 score value is selected as the optimum baseline.

\begin{figure}%
	\centering
	\includegraphics[width=.95\columnwidth]{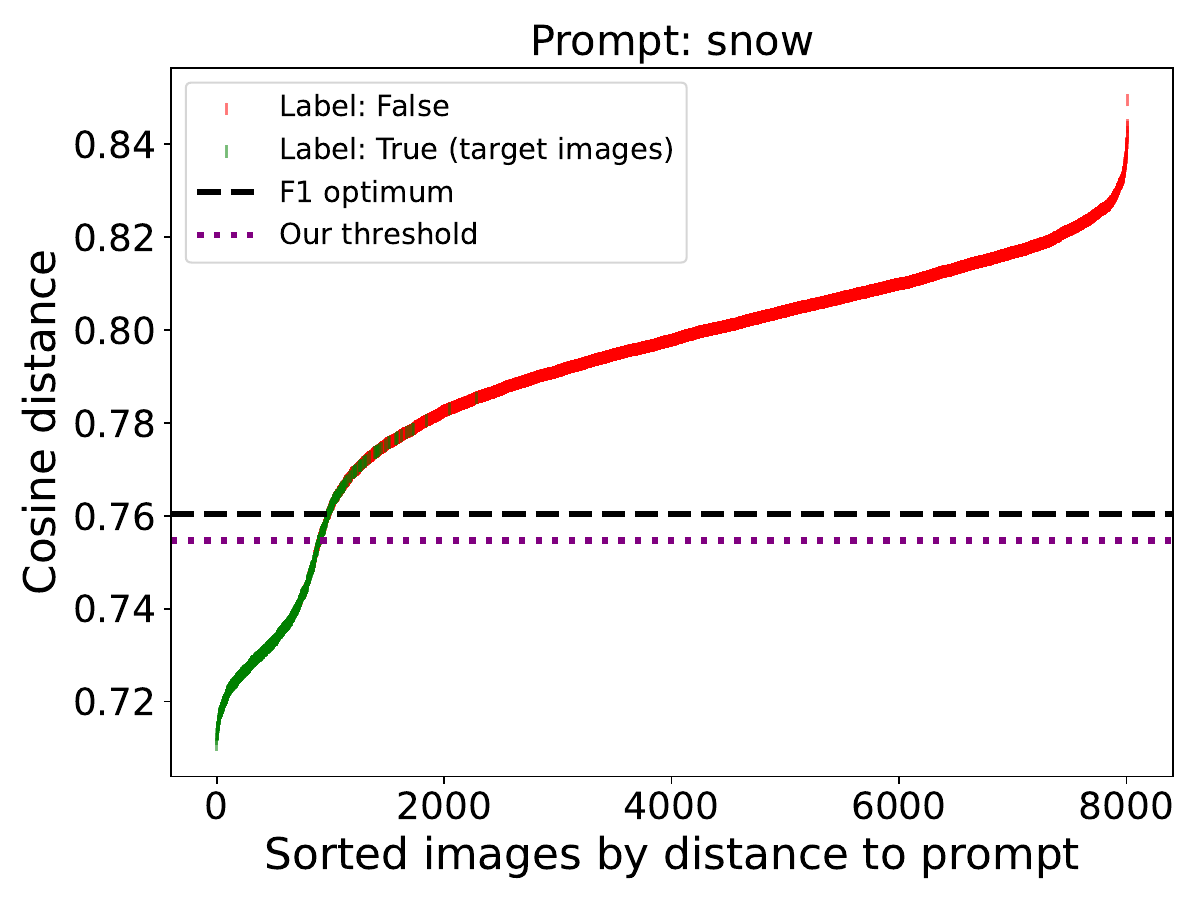}%
	\caption{Images sorted by their cosine distance to the prompt `snow', the threshold based on the optimum F1 score, our threshold, and the ground truth as color coding.}%
	\label{fig:snow_s-curve}%
\end{figure}

\subsection{General Functionality}

First, we arrange the images according to their cosine distance to a certain prompt (see Fig.~\ref{fig:snow_s-curve}).
We can see that CLIP can separate a large part of the images for the `snow' prompt.
However, there is a transition area where the image groups overlap and cannot be differentiated with this prompt.
Of course, we cannot counter this with our threshold value.
However, our threshold value is comparable to the threshold value defined via the optimum F1 score.

\begin{figure}%
	\centering
	\includegraphics[width=.95\columnwidth]{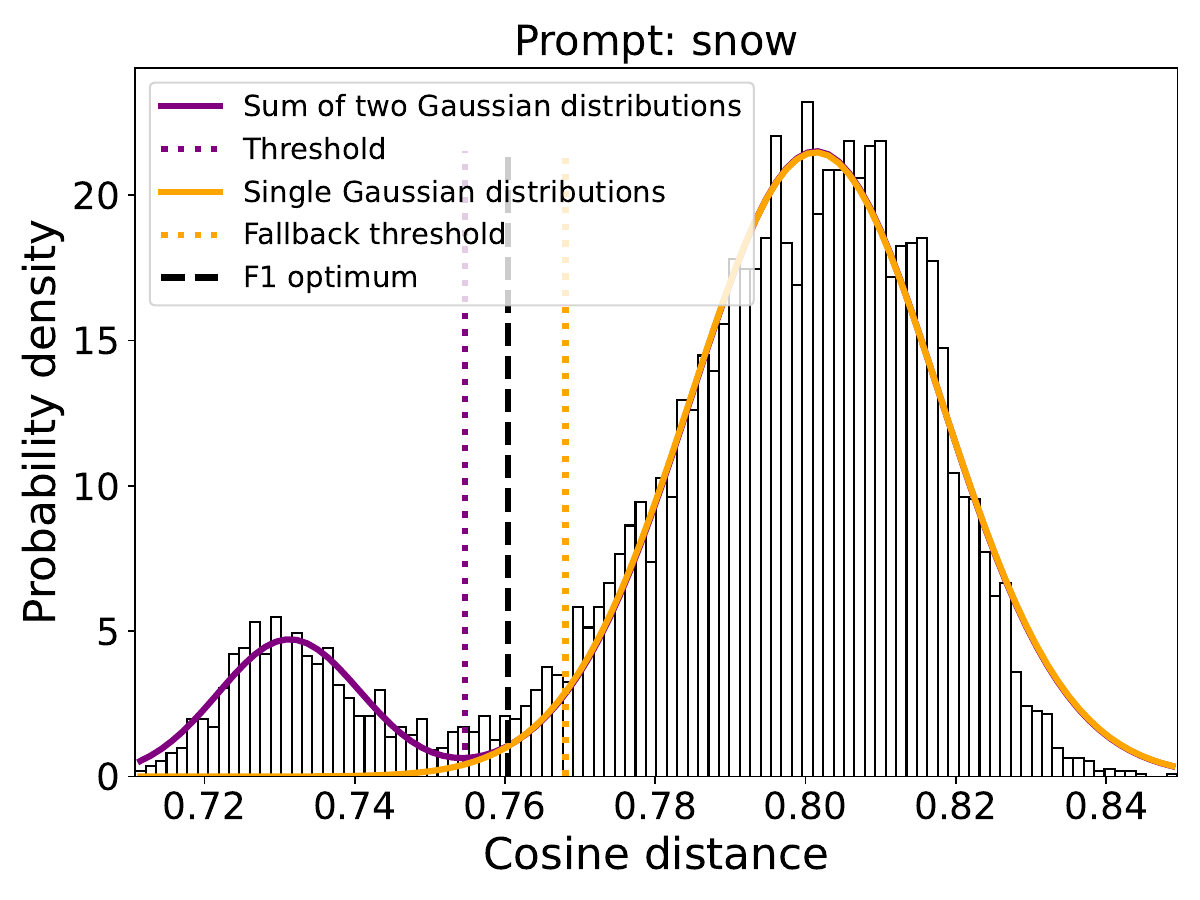}%
	\caption{Distribution of the cosine distances of the ACDC images to the prompt `snow' with the fitted Gaussian distributions and associated thresholds.}%
	\label{fig:snow_histogram}%
\end{figure}

To illustrate our approach, we create the histogram of the cosine distances (see Fig.~\ref{fig:snow_histogram}).
We also show the fitted curves for the sum of two Gaussian distributions and a single Gaussian distribution, including the threshold determined in each case.
For the prompt `snow' the sum of two Gaussian distributions describes the distribution better than the fallback with one single Gaussian distribution.

\begin{figure}
    \centering
    \begin{subfigure}[t]{0.39\textwidth}
        \centering
        \includegraphics[width=\textwidth]{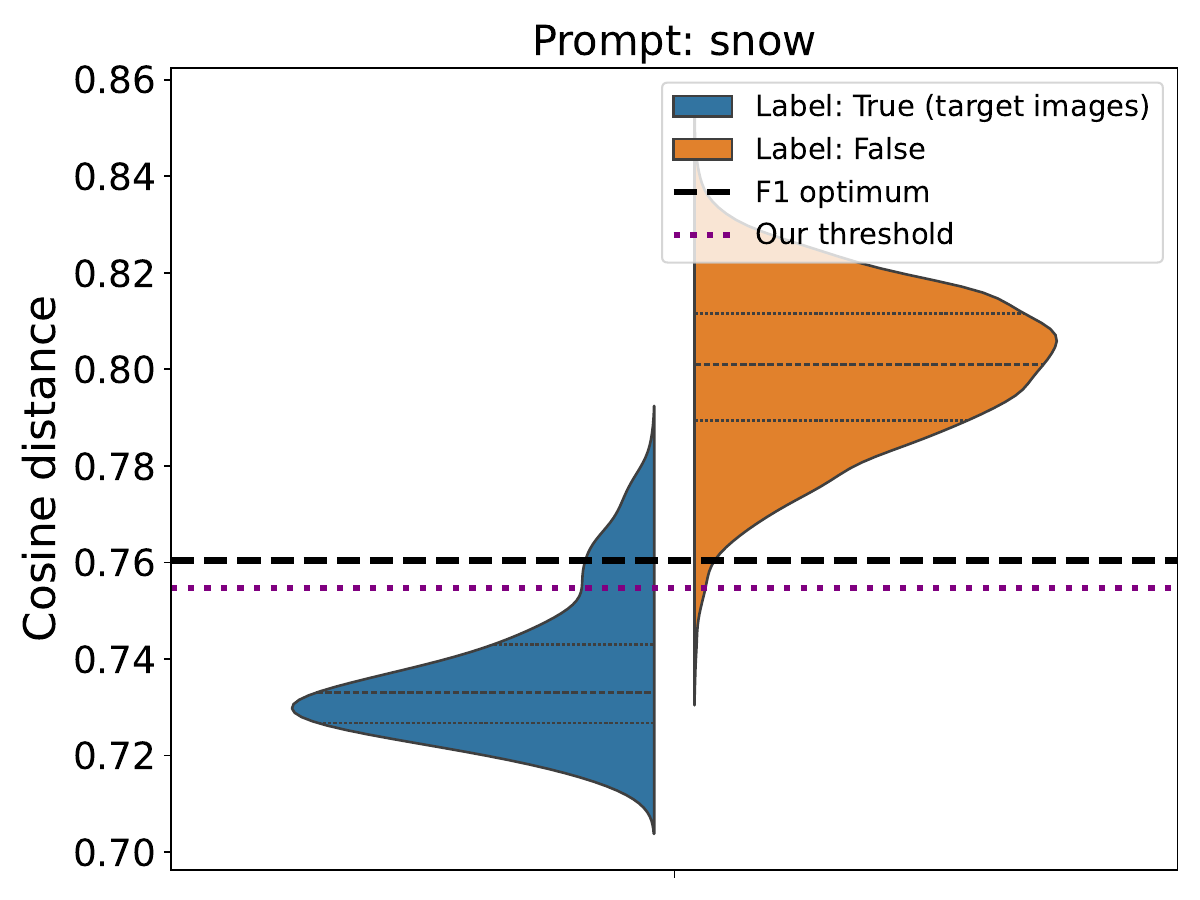}
        \label{fig:violine_plots:a}
    \end{subfigure}
    \hfill
    \begin{subfigure}[t]{0.39\textwidth}
        \centering
        \includegraphics[width=\textwidth]{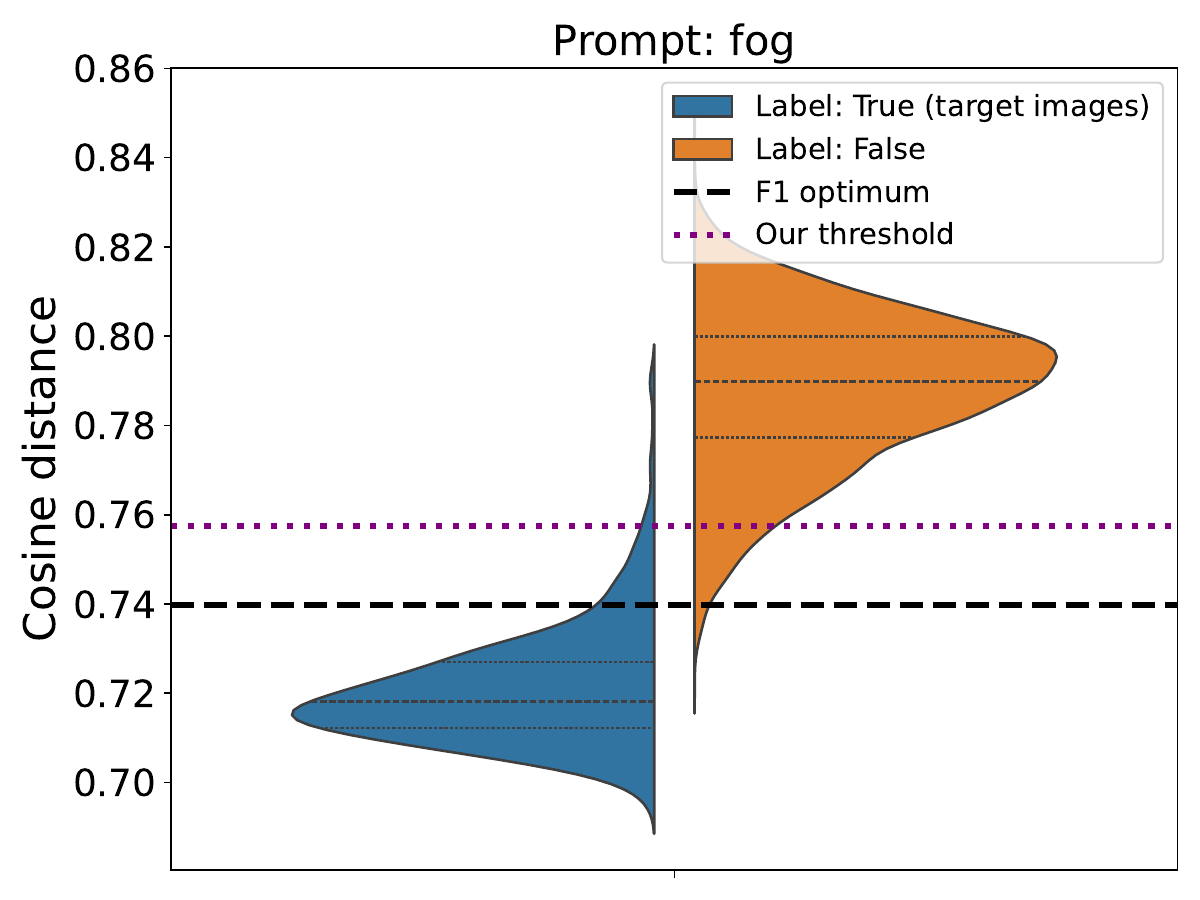}
        \label{fig:violine_plots:b}
    \end{subfigure}
    \hfill
    \begin{subfigure}[t]{0.39\textwidth}
        \centering
        \includegraphics[width=\textwidth]{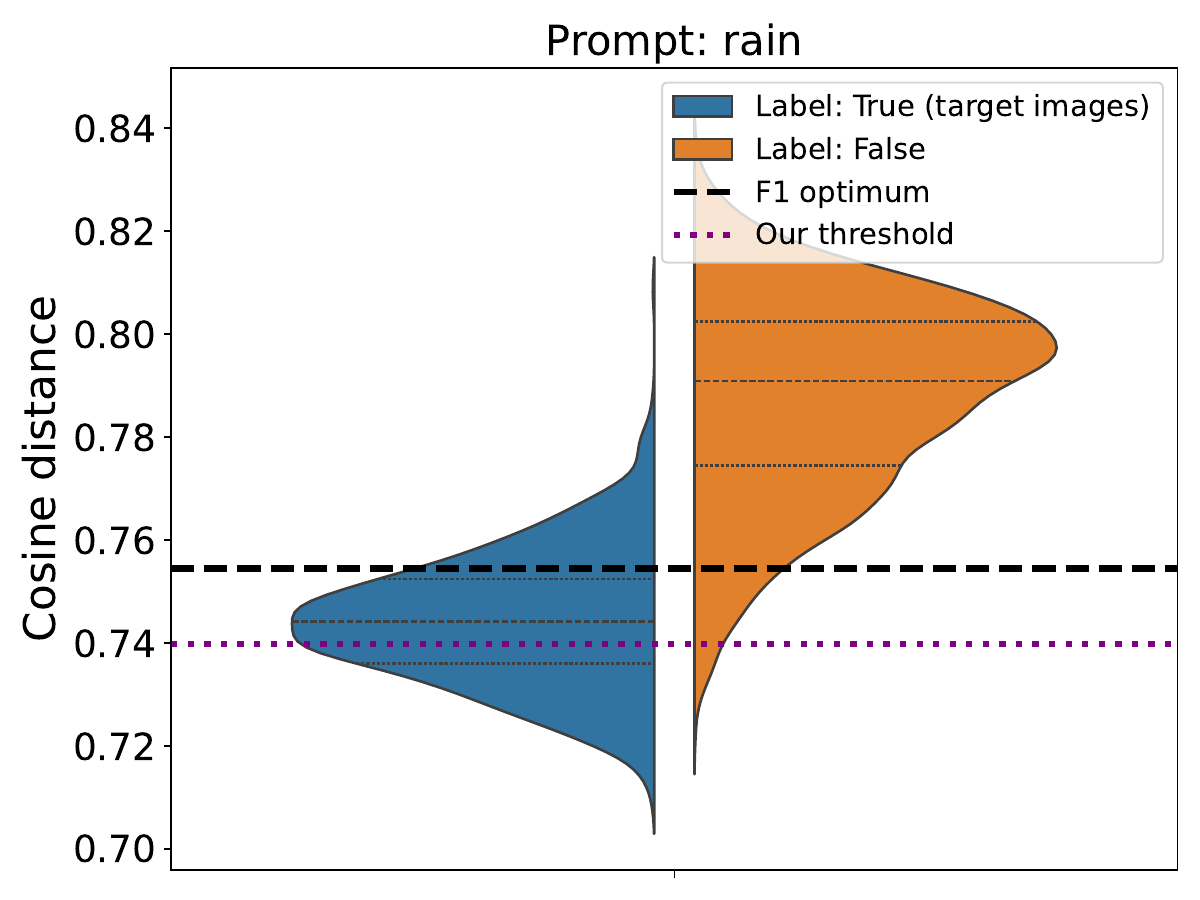}
        \label{fig:violine_plots:c}
    \end{subfigure}
    \hfill
    \begin{subfigure}[t]{0.39\textwidth}
        \centering
        \includegraphics[width=\textwidth]{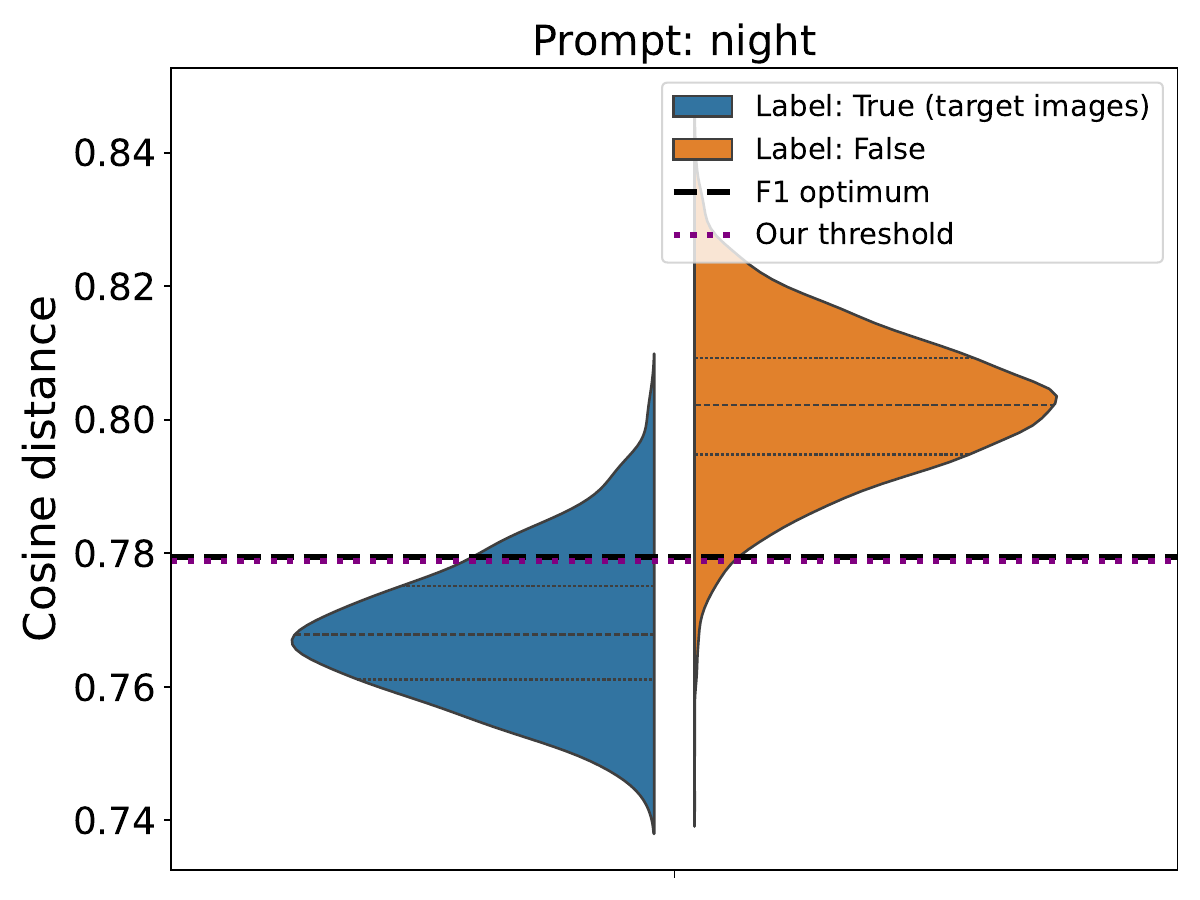}
        \label{fig:violine_plots:d}
    \end{subfigure}
    \caption{Distribution of the cosine distances, separated by their ground truth label, with our threshold, and the threshold based on the optimum F1 score.}%
    \label{fig:violine_plots}
\end{figure} % TODO check
% TODO update size

\begin{table}
	\centering
	\caption{Comparison of our threshold and the optimum F1 score threshold.}%
	\label{tab:comparison_thresholds}%
	\begin{tabular}{lcccc}
		\toprule
		   & \multicolumn{4}{c}{\textbf{Prompt}}\\
          & `snow' & `fog' & `rain' & `night'\\
		\midrule
          \textbf{Our threshold} & & & & \\
          \quad Threshold value & 0.755 & 0.757 & 0.740 & 0.779\\
          \quad Number of results & 902 & 1384 & 442 & 983\\
          \quad F1 score & 0.914 & 0.826 & 0.503 & 0.843\\
          \quad Precision & 0.963 & 0.711 & 0.821 & 0.852\\
          \quad Recall & 0.869 & 0.984 & 0.363 & 0.833\\
          \quad Accuracy & 0.980 & 0.948 & 0.911 & 0.961\\
          \quad Specificity & 0.995 & 0.943 & 0.989 & 0.979\\
          \textbf{Optimum F1 threshold} & & & & \\
          \quad Threshold value & 0.760 & 0.740 & 0.754 & 0.779\\
          \quad Number of results & 979 & 963 & 1232 & 1020\\
          \quad F1 score & 0.924 & 0.945 & 0.729 & 0.844\\
          \quad Precision & 0.934 & 0.963 & 0.660 & 0.838\\
          \quad Recall & 0.914 & 0.927 & 0.813 & 0.850\\
          \quad Accuracy & 0.981 & 0.986 & 0.924 & 0.961\\
          \quad Specificity & 0.991 & 0.995 & 0.940 & 0.976\\
		\bottomrule
	\end{tabular}
\end{table}

\begin{figure*}%
	\centering
	\includegraphics[width=.95\textwidth]{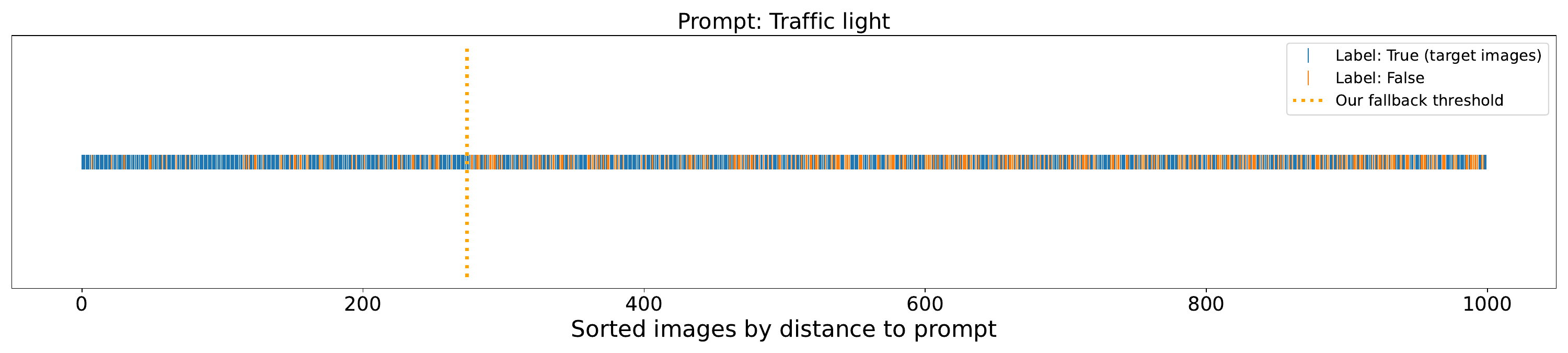}%
	\caption{First 1000 images sorted by their cosine distance to the prompt `traffic light' with our fallback threshold.}%
	\label{fig:traffic_light_barcode}%
\end{figure*}

\begin{figure}%
	\centering
	\includegraphics[width=.95\columnwidth]{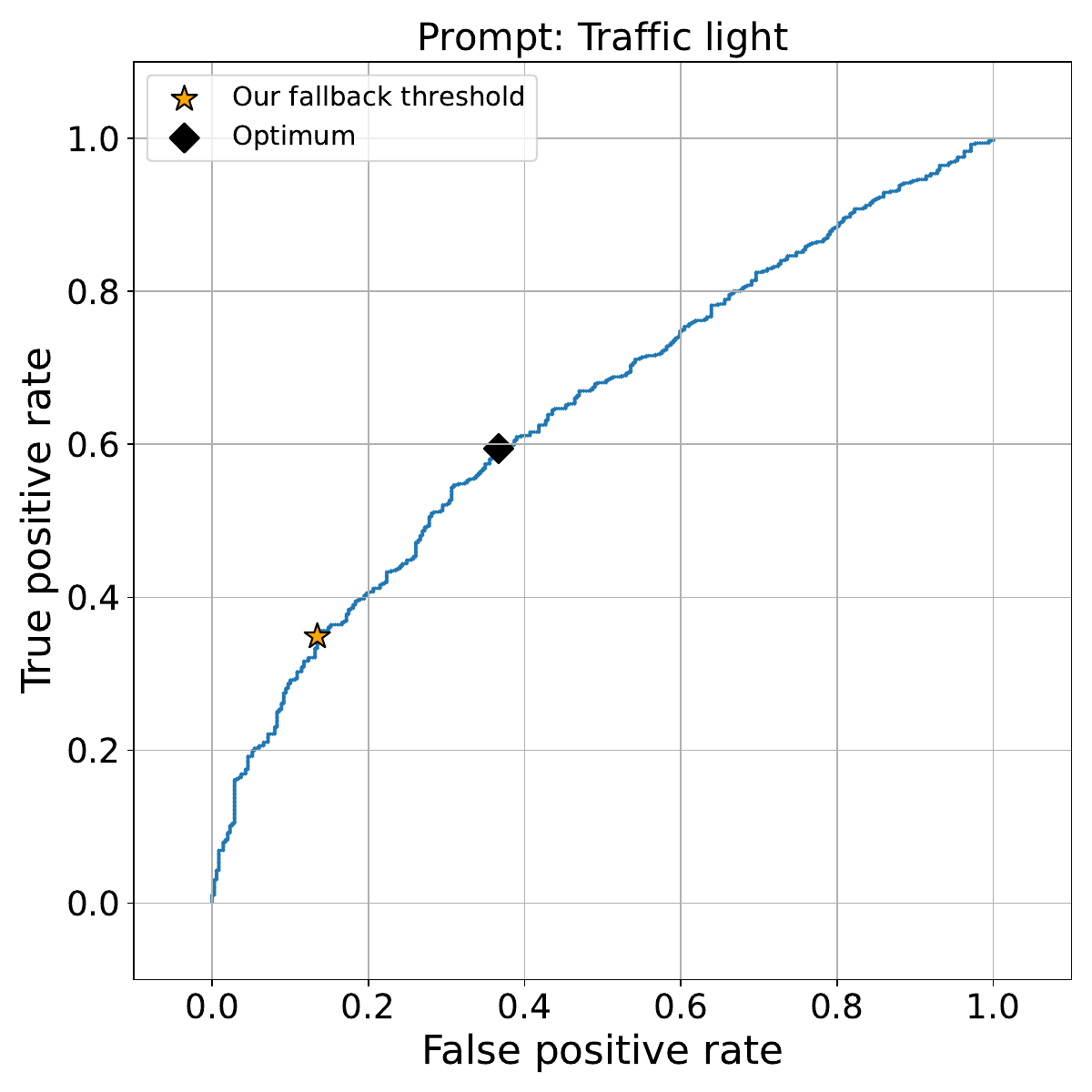}%
	\caption{Receiver operating characteristic curve for the first~1000 labeled for the prompt `traffic light' with our fallback threshold and the optimum.}%
	\label{fig:traffic_light_roc}%
\end{figure}

\subsection{Quantitative Experiments}

We have labels for `snow', `fog', `rain', and `night' in the ACDC data set.
We now look at the extent to which our threshold isolates the matching images for each label~(see Fig.~\ref{fig:violine_plots}).
We use the name of the label as the prompt.

The assumed composition of the distribution of the cosine distances from the images similar to the query~$Q$ and the images dissimilar to the query~$\overline{Q}$ is supported by the visual inspection of the plots.
In fact, for each prompt, the sum of two Gaussian distributions can be used to determine the threshold value.
However, it also becomes clear that it depends on the specific prompt how well the distributions can be split.
Therefore, we compare our threshold to the optimum threshold based on the F1 score on the basis of different classification metrics (see \autoref{tab:comparison_thresholds}).

We find that, as expected by inspection of the plots, the optimum F1 score of~$0.729$ for the prompt `rain' is clearly worse than, for example, for the prompt `fog' with an optimum F1 score of~$0.945$.
This is due to the large overlap between the two distributions.
For this reason, our method also has some difficulties in determining a threshold value.
However,~$\SI{82.1}{\percent}$ of the images~($363$ of $442$) we selected for the prompt `rain' are actually images of the rain class.
On the other hand, we have~$637$ rain images in the total data set that we did not include in the partial data set using our method.
However, in this case we obtain~$363$ images matching our prompt without manual effort, which we could use for developing and testing ADS.

With regard to the proportion of images that do not correspond to the prompt, the results are worst for the prompt `fog', with~$\SI{28.9}{\percent}$ unsuitable images.
For ADS development, this means that~$400$ images have ended up in our partial data set, although we did not want to include them with this prompt.
Conversely, the resulting partial data set contains~$984$ actual images of the fog class.
However, it must be noted that the partial data set also consists of~$1384$ images, which corresponds to~$\SI{12.3}{\percent}$ of the total data set.

In addition, we analyze the size of the resulting partial data sets based on the threshold from the optimum F1 score and the threshold based on our method.
The largest deviation between the baseline and our method can be observed for the label `rain' with a deviation of~$790$ images and the smallest deviation for the label `night' with~$37$ images.

\subsection{Fallback Threshold}
Next, we examine a prompt where our fallback threshold is used.
This is the case for the prompt 'traffic light', as the uncertainty estimate of the parameters for the sum of the two Guassian distributions is more than twice as large as for the individual Gaussian distribution.
We sorted the images based on their cosine distance and labeled the first~1000 images manually.
The following evaluations are based on these labeled images.
Since not all images were labeled, the optimum F1 value cannot be calculated sensibly.
For our fallback threshold we get an F1 score of~$0.491$, a precision of~$0.828$, a recall of~$0.359$, an accuracy of~$0.529$ and a specificity of~$0.866$.
As can also be seen visually, this means that there are many false negatives but only a few false positives (see Fig.~\ref{fig:traffic_light_barcode}).
CLIP makes it possible to sort the images so that there are more images at the beginning that match the prompt than images that do not.
Towards the end of the first~1000 images it is the other way around.
However, two clearly separate classes and a clear boundary can no longer be identified.
This can also be seen from the receiver operating characteristic curve (ROC curve) for the prompt `traffic light' (see Fig.~\ref{fig:traffic_light_roc}).
The rate of true positives and true negatives is plotted for each possible threshold.
The optimum is defined as the point furthest from zero true positives and only false positives.
Even if it is difficult to define a clear boundary, our fallback procedure still finds a threshold value that is in the vicinity of the theoretical optimum.

\section{CONCLUSION}
We have examined a method for creating partial data sets with natural language for the development and testing of ADS.
The starting point is a data set without annotations.
Our method is based on the existing CLIP network, which we use to represent images and texts as vectors.
With the help of CLIP, the images in the data set can be sorted based on similarity to a textual prompt.
However, CLIP only sorts the data set.
To create a partial data record, however, it is necessary to define a threshold value in the sorted list.
Since the method should be used for the development of ADS, this threshold value must be set automatically.
Our developed procedure sets this threshold automatically based on only the result of a single search.
The focus was on achieving a balanced false positive and false negative rate.
A fallback procedure guarantees the automaticity.
All in all, our method reduces the manual effort required to create a partial data set.
This is our contribution to develop and test robust ADS perception systems.

However, the search performance of CLIP is heavily dependent on the specific prompt, which is why the determination of a sensible threshold value also depends on the prompt.
In the future, we therefore want to focus on optimizing the prompt formulation.
In addition, the number of neural networks that identify the semantic of images is constantly increasing.
One of our objectives is to evaluate whether and how these networks are suitable for image retrieval in the automotive context.

\addtolength{\textheight}{-12cm}   % This command serves to balance the column lengths

%%%%%%%%%%%%%%%%%%%%%%%%%%%%%%%%%%%%%%%%%%%%%%%%%%%%%%%%%%%%%%%%%%%%%%%%%%%%%%%%

\section*{ACKNOWLEDGMENT}
This work results from the just better DATA (jbDATA) project supported by the German Federal Ministry for Economic Affairs and Climate Action of Germany (BMWK) and the European Union, grant number 19A23003H.

\bibliographystyle{IEEEtran}
\bibliography{IEEEabrv,root}

\begin{thebibliography}{10}
\providecommand{\url}[1]{#1}
\csname url@rmstyle\endcsname
\providecommand{\newblock}{\relax}
\providecommand{\bibinfo}[2]{#2}
\providecommand\BIBentrySTDinterwordspacing{\spaceskip=0pt\relax}
\providecommand\BIBentryALTinterwordstretchfactor{4}
\providecommand\BIBentryALTinterwordspacing{\spaceskip=\fontdimen2\font plus
\BIBentryALTinterwordstretchfactor\fontdimen3\font minus
  \fontdimen4\font\relax}
\providecommand\BIBforeignlanguage[2]{{%
\expandafter\ifx\csname l@#1\endcsname\relax
\typeout{** WARNING: IEEEtran.bst: No hyphenation pattern has been}%
\typeout{** loaded for the language `#1'. Using the pattern for}%
\typeout{** the default language instead.}%
\else
\language=\csname l@#1\endcsname
\fi
#2}}

\bibitem{radford_learning_2021-1}
\BIBentryALTinterwordspacing
A.~Radford, J.~W. Kim, C.~Hallacy, A.~Ramesh, G.~Goh, S.~Agarwal, G.~Sastry,
  A.~Askell, P.~Mishkin, J.~Clark, G.~Krueger, and I.~Sutskever, ``Learning
  {{Transferable Visual Models From Natural Language Supervision}},'' Feb.
  2021. [Online]. Available: \url{http://arxiv.org/abs/2103.00020}
\BIBentrySTDinterwordspacing

\bibitem{xia_automated_2023}
\BIBentryALTinterwordspacing
X.~Xia, Z.~Meng, X.~Han, H.~Li, T.~Tsukiji, R.~Xu, Z.~Zheng, and J.~Ma, ``An
  automated driving systems data acquisition and analytics platform,''
  \emph{Transportation Research Part C: Emerging Technologies}, vol. 151, p.
  104120, June 2023. [Online]. Available:
  \url{https://linkinghub.elsevier.com/retrieve/pii/S0968090X23001092}
\BIBentrySTDinterwordspacing

\bibitem{king_taxonomy_2020}
\BIBentryALTinterwordspacing
C.~King, L.~Ries, J.~Langner, and E.~Sax, ``A {{Taxonomy}} and {{Survey}} on
  {{Validation Approaches}} for {{Automated Driving Systems}},'' in \emph{2020
  {{IEEE International Symposium}} on {{Systems Engineering}}
  ({{ISSE}})}.\hskip 1em plus 0.5em minus 0.4em\relax Vienna, Austria: IEEE,
  Oct. 2020, pp. 1--8. [Online]. Available:
  \url{https://ieeexplore.ieee.org/document/9272219/}
\BIBentrySTDinterwordspacing

\bibitem{liu_survey_2024}
\BIBentryALTinterwordspacing
M.~Liu, E.~Yurtsever, X.~Zhou, J.~Fossaert, Y.~Cui, B.~L. Zagar, and A.~C.
  Knoll, ``A {{Survey}} on {{Autonomous Driving Datasets}}: {{Data Statistic}},
  {{Annotation}}, and {{Outlook}},'' Jan. 2024. [Online]. Available:
  \url{http://arxiv.org/abs/2401.01454}
\BIBentrySTDinterwordspacing

\bibitem{alkhawlani_text-based_2015}
M.~Alkhawlani and M.~Elmogy, ``Text-based, {{Content-based}}, and
  {{Semantic-based Image Retrievals}}: {{A Survey}},'' vol.~04, no.~01, 2015.

\bibitem{stefanini_show_2023}
\BIBentryALTinterwordspacing
M.~Stefanini, M.~Cornia, L.~Baraldi, S.~Cascianelli, G.~Fiameni, and
  R.~Cucchiara, ``From {{Show}} to {{Tell}}: {{A Survey}} on {{Deep
  Learning-Based Image Captioning}},'' \emph{IEEE Transactions on Pattern
  Analysis and Machine Intelligence}, vol.~45, no.~1, pp. 539--559, Jan. 2023.
  [Online]. Available: \url{https://ieeexplore.ieee.org/document/9706348/}
\BIBentrySTDinterwordspacing

\bibitem{naito_browsing_2010}
\BIBentryALTinterwordspacing
M.~Naito, C.~Miyajima, T.~Nishino, N.~Kitaoka, and K.~Takeda, ``A browsing and
  retrieval system for driving data,'' in \emph{2010 {{IEEE Intelligent
  Vehicles Symposium}}}.\hskip 1em plus 0.5em minus 0.4em\relax La Jolla, CA,
  USA: IEEE, June 2010, pp. 1159--1165. [Online]. Available:
  \url{http://ieeexplore.ieee.org/document/5547999/}
\BIBentrySTDinterwordspacing

\bibitem{klitzke_real-world_2019}
\BIBentryALTinterwordspacing
L.~Klitzke, C.~Koch, A.~Haja, and F.~K{\"o}ster, ``Real-world {{Test Drive
  Vehicle Data Management System}} for {{Validation}} of {{Automated Driving
  Systems}}:,'' in \emph{Proceedings of the 5th {{International Conference}} on
  {{Vehicle Technology}} and {{Intelligent Transport Systems}}}.\hskip 1em plus
  0.5em minus 0.4em\relax Heraklion, Crete, Greece: {SCITEPRESS - Science and
  Technology Publications}, 2019, pp. 171--180. [Online]. Available:
  \url{https://www.scitepress.org/DigitalLibrary/Link.aspx?doi=10.5220/0007720501710180}
\BIBentrySTDinterwordspacing

\bibitem{rigoll_scalable_2022}
\BIBentryALTinterwordspacing
P.~Rigoll, L.~Ries, and E.~Sax, ``Scalable {{Data Set Distillation}} for the
  {{Development}} of {{Automated Driving Functions}},'' in \emph{2022 {{IEEE}}
  25th {{International Conference}} on {{Intelligent Transportation Systems}}
  ({{ITSC}})}.\hskip 1em plus 0.5em minus 0.4em\relax Macau, China: IEEE, Oct.
  2022, pp. 3139--3145. [Online]. Available:
  \url{https://ieeexplore.ieee.org/document/9921868/}
\BIBentrySTDinterwordspacing

\bibitem{lew_content-based_2006}
\BIBentryALTinterwordspacing
M.~S. Lew, N.~Sebe, C.~Djeraba, and R.~Jain, ``Content-based multimedia
  information retrieval: {{State}} of the art and challenges,'' \emph{ACM
  Transactions on Multimedia Computing, Communications, and Applications},
  vol.~2, no.~1, pp. 1--19, Feb. 2006. [Online]. Available:
  \url{https://dl.acm.org/doi/10.1145/1126004.1126005}
\BIBentrySTDinterwordspacing

\bibitem{rigoll_focus_2023}
\BIBentryALTinterwordspacing
P.~Rigoll, P.~Petersen, H.~Stage, L.~Ries, and E.~Sax, ``Focus on the
  {{Challenges}}: {{Analysis}} of a {{User-friendly Data Search Approach}} with
  {{CLIP}} in the {{Automotive Domain}},'' in \emph{2023 {{IEEE}} 26th
  {{International Conference}} on {{Intelligent Transportation Systems}}
  ({{ITSC}})}.\hskip 1em plus 0.5em minus 0.4em\relax Bilbao, Spain: IEEE,
  Sept. 2023, pp. 168--174. [Online]. Available:
  \url{https://ieeexplore.ieee.org/document/10422271/}
\BIBentrySTDinterwordspacing

\bibitem{rigoll_unveiling_2023}
\BIBentryALTinterwordspacing
P.~Rigoll, J.~Langner, and E.~Sax, ``Unveiling {{Objects}} with {{SOLA}}: {{An
  Annotation-Free Image Search}} on the {{Object Level}} for {{Automotive Data
  Sets}},'' Dec. 2023. [Online]. Available:
  \url{http://arxiv.org/abs/2312.01860}
\BIBentrySTDinterwordspacing

\bibitem{national_institute_of_standards_and_technology_nist_cosine_2023}
\BIBentryALTinterwordspacing
{National Institute of Standards and Technology (NIST)}, ``{{COSINE DISTANCE}},
  {{COSINE SIMILARITY}}, {{ANGULAR COSINE DISTANCE}}, {{ANGULAR COSINE
  SIMILARITY}},'' Mar. 2023. [Online]. Available:
  \url{https://www.itl.nist.gov/div898/software/dataplot/refman2/auxillar/cosdist.htm}
\BIBentrySTDinterwordspacing

\bibitem{sakaridis_acdc_2021}
\BIBentryALTinterwordspacing
C.~Sakaridis, D.~Dai, and L.~Van~Gool, ``{{ACDC}}: {{The Adverse Conditions
  Dataset}} with {{Correspondences}} for {{Semantic Driving Scene
  Understanding}},'' in \emph{2021 {{IEEE}}/{{CVF International Conference}} on
  {{Computer Vision}} ({{ICCV}})}.\hskip 1em plus 0.5em minus 0.4em\relax
  Montreal, QC, Canada: IEEE, Oct. 2021, pp. 10\,745--10\,755. [Online].
  Available: \url{https://ieeexplore.ieee.org/document/9711067/}
\BIBentrySTDinterwordspacing

\bibitem{wang_balanced_2024}
\BIBentryALTinterwordspacing
H.~Wang, Y.~Zhan, L.~Liu, L.~Ding, and J.~Yu, ``Balanced {{Similarity}} with
  {{Auxiliary Prompts}}: {{Towards Alleviating Text-to-Image Retrieval Bias}}
  for {{CLIP}} in {{Zero-shot Learning}},'' Feb. 2024. [Online]. Available:
  \url{http://arxiv.org/abs/2402.18400}
\BIBentrySTDinterwordspacing

\end{thebibliography}

\end{document}